\documentclass[11pt]{article}

\usepackage[utf8]{inputenc}  
\usepackage[T1]{fontenc}     
\usepackage{amsmath, amssymb} 
\usepackage{hyperref}        
\usepackage{url}             
\usepackage{booktabs}        
\usepackage{amsfonts}        
\usepackage{nicefrac}        
\usepackage{microtype}       
\usepackage{xcolor}          
\usepackage[numbers]{natbib}
\usepackage{graphicx} 
\usepackage{float}    
\usepackage{subcaption}

\usepackage{algorithm}
\usepackage{algpseudocode}

\usepackage{geometry}
\title{\bfseries P-DROP: Poisson-Based Dropout for Graph Neural Networks
}
\author{
  Hyunsik Yun \\
  School of Mathematical and Statistical Sciences\\
  Arizona State University\\
  \texttt{hyun26@asu.edu}
}
\date{}

\begin{document}

\maketitle

\begin{abstract}
Over-smoothing remains a major challenge in Graph Neural Networks (GNNs), where repeated message passing causes node representations to converge and lose discriminative power. To address this, we propose a novel node selection strategy based on Poisson processes, introducing stochastic but structure-aware updates. Specifically, we equip each node with an independent Poisson clock, enabling asynchronous and localized updates that preserve structural diversity. We explore two applications of this strategy: as a replacement for dropout-based regularization and as a dynamic subgraph training scheme. Experimental results on standard benchmarks (Cora, Citeseer, Pubmed) demonstrate that our Poisson-based method yields competitive or improved accuracy compared to traditional Dropout, DropEdge, and DropNode approaches, particularly in later training stages.

\end{abstract}

\section{Introduction}
Graphs are widely used in various applications due to their ability to model relationships between entities (nodes) through connectivity. This structural information allows for more expressive representations of complex datasets.

With the rise of machine learning, Graph Neural Networks (GNNs) have gained popularity. Along with this rise, several challenges have emerged. One of the most well-known issues is over-smoothing. This problem arises from the very mechanism that gives GNNs their strength—message passing along edges. Most GNN architectures aggregate information from a node’s local neighborhood to capture the graph's structure. However, as more GNN layers are stacked, node representations tend to become increasingly similar, eventually converging to indistinguishable vectors. Unlike overfitting, which occurs when a model fits the training data too closely and fails to generalize to unseen data, over-smoothing is a structural problem that results from excessive propagation across the graph.

To mitigate over-smoothing, various sampling and propagation control strategies have been proposed (\citet{rusch2023surveyoversmoothinggraphneural}.) In this paper, we introduce a node selection mechanism based on a Poisson process, which probabilistically controls the propagation path in GNNs by leveraging local connectivity patterns.

\section{Background}
\subsection{Poisson-Based Update Strategy}

We introduce a stochastic update mechanism for graph neural networks based on the theory of Poisson processes. This framework provides a principled way to regulate update sparsity and timing, which helps reduce over-smoothing and supports localized information propagation.

A Poisson process is a fundamental stochastic model for random events occurring over time with a constant average rate $\lambda > 0$. Formally, let $\{N(t): t \geq 0\}$ be a Poisson process with rate $\lambda$. The number of events in any interval of length $t$ follows a Poisson distribution:
\[
\mathbb{P}(N(t) = k) = \frac{(\lambda t)^k}{k!} e^{-\lambda t}, \quad k = 0, 1, 2, \dots
\]
A key property is that the inter-arrival times between consecutive events are i.i.d.\ exponential random variables with rate $\lambda$:
\[
T \sim \mathrm{Exp}(\lambda), \quad f_T(t) = \lambda e^{-\lambda t}, \quad t \geq 0.
\]

\paragraph{Local Clocks and Asynchronous Updates.}
In our method, each node $v \in V$ is equipped with an independent Poisson clock with rate $\lambda_v$, which determines when it becomes eligible for update. At each time $t$, we define the active node set as:
\[
V_{\text{active}}(t) = \{v \in V : T_v \leq t\}.
\]
This asynchronous update scheme ensures that:
\begin{itemize}
    \item Nodes update sparsely and independently, mitigating simultaneous over-updating.
    \item Node update frequency can be customized via $\lambda_v$, reflecting structural importance.
\end{itemize}
The exponential distribution’s memoryless property further suits recurrent and decentralized update scheduling.

\paragraph{Superposition and Indirect Influence.}
The superposition property of Poisson processes states that the union of independent Poisson processes is itself a Poisson process. As a result, although a node may not be directly active at time $t$, it may still participate indirectly in the update through neighbors whose clocks have triggered. The aggregated activity ensures that nodes embedded in highly connected or frequently updated regions continue to receive information, even if they are not directly selected.

\paragraph{Comparison to Uniform Dropout Methods.}
Traditional methods such as DropEdge or DropMessage randomly suppress nodes or edges uniformly, regardless of graph structure. This may inadvertently remove crucial hubs or disconnect important subgraphs, degrading message propagation. In contrast, our Poisson-based strategy preserves structural integrity, as key nodes either update directly (via high $\lambda_v$) or remain involved through their neighbors.

\paragraph{Relation to Stochastic Models.}
Our approach is inspired by stochastic interacting particle systems such as the contact process (\citet{lanchier2024stochastic}), where local Poisson clocks drive the spread of information across space and time. Similarly, in our setting, each node initiates local propagation based on its own stochastic clock, enabling spatially localized and temporally diverse updates.

\begin{figure}[H]
    \centering
    \begin{minipage}{0.32\textwidth}
        \centering
        \includegraphics[width=\linewidth]{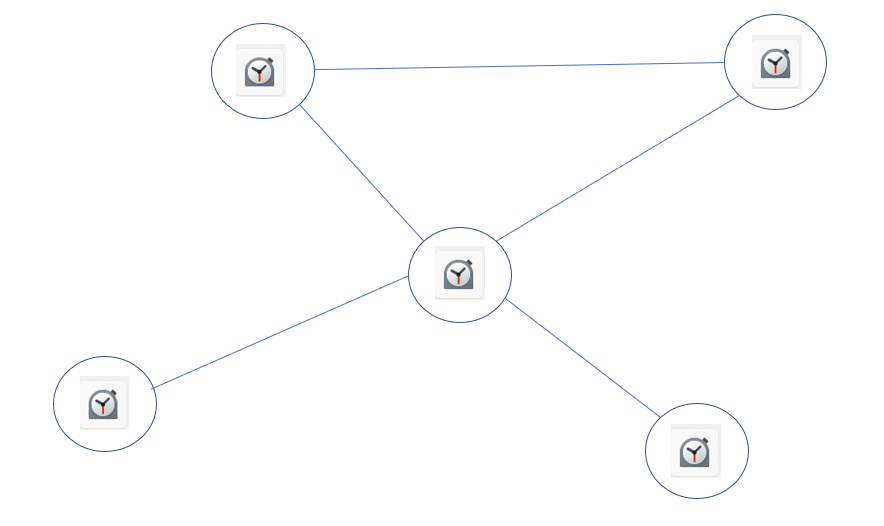}
        \caption{Graph}
    \end{minipage}
    \hfill
    \begin{minipage}{0.32\textwidth}
        \centering
        \includegraphics[width=\linewidth]{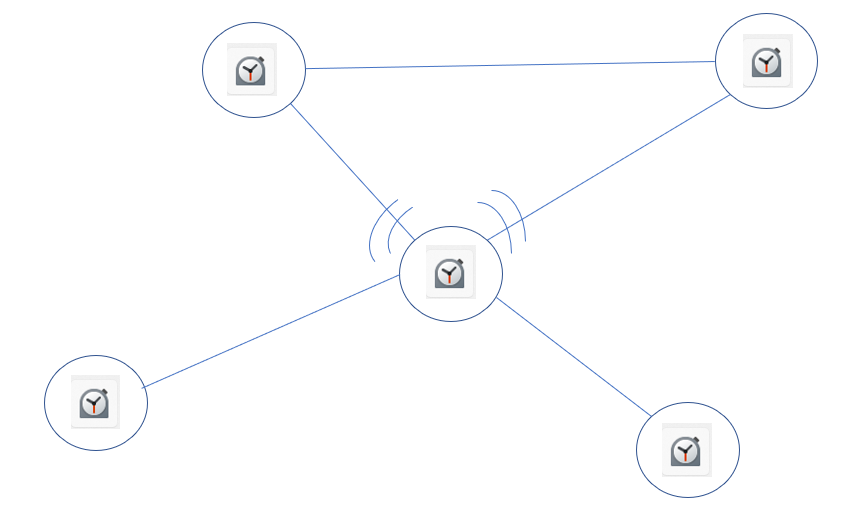}
        \caption{One Alarm rings}
    \end{minipage}
    \hfill
    \begin{minipage}{0.32\textwidth}
        \centering
        \includegraphics[width=\linewidth]{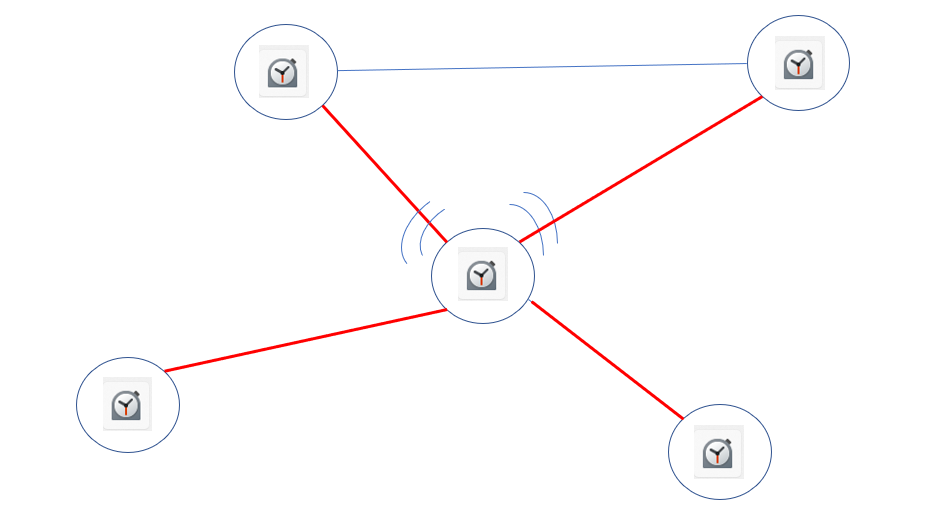}
        \caption{propogate neighborhoods}
    \end{minipage}
    \caption*{How it works}
\end{figure}

\subsection{Sampling-Based Approaches for Over-Smoothing}

As Graph Neural Networks (GNNs) aggregate feature information across neighboring nodes through multiple layers, a well-known problem called over-smoothing arises. In deep GNNs, repeated message passing causes node representations to become indistinguishably similar, ultimately degrading performance. To alleviate this issue, several sampling-based strategies have been proposed that reduce the extent of propagation or the amount of information exchanged among nodes.

For GNN, it uses sampling like Dropout (\citet{JMLR:v15:srivastava14a}), however, it utilizes structures of GNN.
DropEdge (\citet{Rong2020DropEdge:}) is one of the earliest and most widely used techniques in this category. It randomly removes a subset of edges in the input graph during training, thereby weakening the structural connectivity and limiting the influence radius of each node. This helps maintain diversity in node representations across layers.

More recently, DropMessage (\citet{Fang_Xiao_Wang_Xu_Yang_Yang_2023}) introduces stochasticity at the message-passing level. Rather than modifying the graph structure itself, it randomly skips messages passed along existing edges. This fine-grained form of regularization reduces redundancy in aggregated information while preserving the overall topology.

While these methods have shown effectiveness in mitigating over-smoothing, most of them rely on uniform random sampling, which may not account for the underlying structure of the graph. In this work, we aim to address this limitation by introducing a Poisson process-based node selection mechanism that leverages local connectivity patterns to probabilistically control propagation paths.

\subsection{Suitability of the Method for Graph-Structured Data}

\begin{figure}[H]
     \centering
     \includegraphics[width=0.6\linewidth]{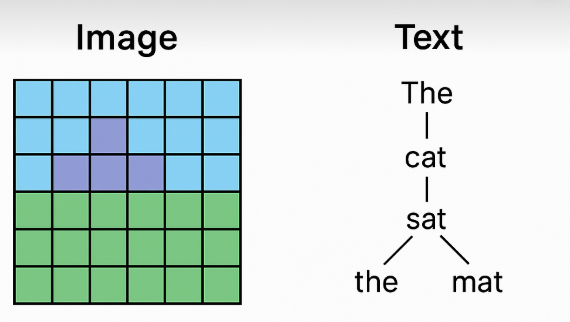}
        \caption{Illustration of other data : Image, Text}
\end{figure}

While the proposed subgraph update scheme works well for graphs, it does not naturally extend to other data types such as images or text. These data are typically structured and dense: images rely on convolutional architectures (\citet{oshea2015introductionconvolutionalneuralnetworks}) that exploit spatial locality, while language models are commonly built upon sequential architectures like Recurrent Neural Networks (\citet{elman1990finding}) or Long Short-Term Memory networks (\citet{10.1162/neco.1997.9.8.1735}). In such domains, there is no inherent notion of sparse, evolving subgraphs, making our method less applicable.

\begin{figure}[H]
     \centering
     \includegraphics[width=0.6\linewidth]{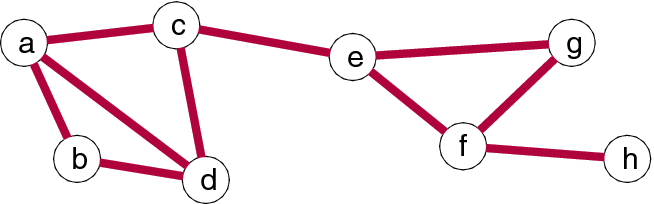}
        \caption{Illustration of node degree imbalance in graph structures}
\end{figure}

Graph-structured data is inherently non-uniform: nodes have varying degrees, and those with higher degrees tend to play a more significant role in message propagation. Traditional update schemes often ignore this imbalance by treating all nodes equally. In contrast, our method leverages the natural variability in node connectivity through Poisson-based subgraph updates, allowing structurally important nodes to influence learning more frequently.

\section{Proposed Methods}

Our proposed idea can be naturally extended in two distinct directions, both grounded in the use of Poisson processes for node selection. The first direction is to reinterpret dropout in Graph Neural Networks through a Poisson-based mechanism. Traditional dropout methods, such as DropNode or DropEdge, rely on fixed probabilities or heuristics to randomly remove nodes or edges during training. In contrast, our approach introduces a stochastic but structurally-aware alternative by sampling nodes according to a Poisson process, potentially making dropout more adaptive to the underlying graph structure.

The second direction involves using Poisson-based sampling not merely as a regularization technique, but as a core component of the model's update scheme. Specifically, rather than operating on the entire graph at each time step, we propose updating the model over dynamically selected subgraphs, where the selection is governed by a Poisson process. This perspective reframes training as a sequence of local updates on sampled regions of the graph, potentially reducing over-smoothing and computational costs while still preserving structural expressiveness.

We explore both of these directions in the subsequent subsections.

\subsection{Poisson-Based Alternatives to Dropout Regularization in GNNs}

As previously discussed, one common approach to mitigating overfitting and over-smoothing in GNNs is to apply dropout-based regularization techniques.

\begin{algorithm}
\caption{Poisson-Based Node Selection for Dropout in GNN}
\begin{algorithmic}[1]
\State \textbf{Input:} Graph \( G = (V, E) \), intensity \( \lambda \), cutoff threshold \( t_{\mathrm{cut}} \)
\State \textbf{Sample:} For each node \( v \in V \), draw \( T_v \sim \mathrm{Exp}(\lambda) \)
\State \textbf{Select nodes:} \( V_{\mathrm{active}} \gets \{ v \in V \mid T_v \leq t_{\mathrm{cut}} \} \)
\State \textbf{Induce subgraph:} \( G_{\mathrm{active}} = (V_{\mathrm{active}}, E_{\mathrm{active}}) \)
\State \textbf{Run GNN:} Perform forward and backward propagation only on \( G_{\mathrm{active}} \)
\end{algorithmic}
\end{algorithm}

In the proposed algorithm, each node is assigned an independent exponential random variable with rate $\lambda$. Nodes whose sampled values greater than the threshold are selected to form the subgraph used for propagation.

This Poisson-based node selection scheme can be interpreted as a structural alternative to standard Dropout. Unlike uniform random sampling, it allows the selection process to be sensitive to the structural properties of the graph, thereby introducing a form of structure-aware regularization.

\subsection{Poisson-Based Dynamic Subgraph Updates for GNN Training}

\begin{algorithm}
\caption{Poisson-Based Dynamic Subgraph Updates for GNN Training}
\begin{algorithmic}[1]
\State \textbf{Input:} Graph \( G = (V, E) \), intensity \( \lambda \), total time \( T \), step size \( \Delta t \)
\State \textbf{Initialize:} For each node \( v \in V \), sample \( T_v \sim \mathrm{Exp}(\lambda) \)
\State Set current time \( t \gets 0 \)
\While{ \( t < T \) }
    \State \( V_t \gets \{ v \in V \mid t \geq T_v \} \)
    \State Induce subgraph \( G_t = (V_t, E_t) \) from \( G \)
    \State Run GNN forward/backward propagation on \( G_t \)
    \For{each \( v \in V_t \)}
        \State Resample: \( T_v \gets T_v + \mathrm{Exp}(\lambda) \)
    \EndFor
    \State \( t \gets t + \Delta t \)
\EndWhile
\end{algorithmic}
\end{algorithm}

\section{Experiments}

We evaluate the effectiveness of our Poisson-based update strategy through a series of controlled experiments on three standard citation network benchmarks: \textbf{Cora}, \textbf{CiteSeer}, and \textbf{PubMed}. Our evaluation is divided into two parts. First, we compare our method against standard dropout-based regularization strategies. Then, we assess its performance in the context of subgraph-based propagation, contrasting it with other scalable and localized GNN training paradigms.

\subsection{Datasets}

We use three widely adopted citation network datasets:

\begin{itemize}
    \item \textbf{Cora}: 2,708 nodes, 5,429 edges, 7 classes.
    \item \textbf{CiteSeer}: 3,327 nodes, 4,732 edges, 6 classes.
    \item \textbf{PubMed}: 19,717 nodes, 44,338 edges, 3 classes.
\end{itemize}
Each node represents a scientific publication, and edges denote citation links. Node features are bag-of-words representations, and the task is to predict the research category of each node.

\section{Results}

\subsection{Drop-Based Methods}

We used PyTorch Geometric to implement various dropout-based regularization methods. However, DropMessage was not compatible with the current library version, and due to implementation issues—such as Python version incompatibility—we were unable to include it in our experiments.

We used the name for our method SGNN(Stochastic Graph Neural Networks).

The full validation and test accuracy curves for 200 training epochs are provided in Chapter 8, Appendix.

\begin{table}[h]
\centering
\resizebox{\textwidth}{!}{%
\begin{tabular}{lcccccc}
\toprule
\textbf{Method} &
\textbf{Cora Val} & \textbf{Cora Test} &
\textbf{Citeseer Val} & \textbf{Citeseer Test} &
\textbf{Pubmed Val} & \textbf{Pubmed Test} \\
\midrule
Dropout   & \textbf{0.7800} & 0.7940 & \textbf{0.6820} & 0.6720 & 0.7880 & 0.7900 \\
DropEdge  & 0.7660 & 0.8110 & 0.6760 & \textbf{0.6830} & 0.7760 & 0.7880 \\
DropNode  & 0.7760 & 0.8050 & 0.6720 & 0.6810 & 0.7860 & 0.7870 \\
SGNN      & 0.7720 & \textbf{0.8120} & 0.6600 & 0.6620 & \textbf{0.7940} & \textbf{0.7930} \\
\bottomrule
\end{tabular}
}

\caption{Final validation and test accuracy on Cora, Citeseer, and Pubmed.}
\label{tab:comparison-all}
\end{table}

Table 1 summarizes the final validation and test accuracy for each method across the three benchmark datasets. Overall, the performance differences among the models are relatively small. However, our proposed SGNN method achieved highly competitive results. In particular, SGNN obtained the highest test accuracy on Pubmed and matched or exceeded the performance of DropEdge and DropNode on Cora. While Citeseer results were slightly lower, the model still remained within a close margin. These results indicate that the Poisson-based update mechanism is effective and robust across different graph structures.

\section{Conclusion}

In this project, we introduced a Poisson-based node selection mechanism for Graph Neural Networks to address the over-smoothing problem. By assigning each node an independent Poisson clock, our method enables asynchronous and structure-aware updates, either as a dropout regularization technique or as a subgraph-based training framework. Experiments on standard citation networks show that, despite slower initial performance, our method converges to comparable or superior accuracy in later stages, especially on larger graphs like Pubmed. These results suggest that leveraging stochastic update timing informed by graph structure can offer a promising direction for improving training efficiency and performance in GNNs.

\section{Future Work}

\subsection{Over-Smoothing}

The initial motivation for this model was to address the over-smoothing problem inherent in deep Graph Neural Networks. While the theoretical framework has been established, the current implementation is still under development due to limited time and computational resources. To advance this direction, we plan to analyze and build upon existing implementations from prior studies, which have explored various strategies to mitigate over-smoothing.

In particular, we observed promising results from our drop-based models, suggesting that the Poisson-based update mechanism may also serve as an effective regularization tool against over-smoothing. These preliminary findings motivate further exploration and refinement of the method, especially in deeper architectures where the issue becomes more pronounced.

\subsection{Computation Cost Analysis}

At first glance, the use of exponential random variables for scheduling node updates may seem computationally intensive, especially when applied across all nodes in a large graph. It may seem computationally expensive to sample a random variable for each node and compare it during each update, especially for large graphs.

 However, in practice, the process reduces to assigning a single sampled value to each node and performing a simple comparison with a threshold or current time. This results in an overall computational complexity of approximately $O(N),$ where $N$ is the total number of nodes. Since the exponential sampling can be efficiently vectorized and comparisons are element-wise, the method scales well in typical GPU-accelerated environments. Nonetheless, a more rigorous analysis and empirical benchmarking are needed to fully quantify the trade-offs involved, particularly in dynamic subgraph settings or when varying the Poisson rates across nodes.

\subsection{Comparison in Subgraph Propagation Settings}
We are currently in the process of extending our Poisson-based update mechanism to subgraph propagation settings, which are commonly employed to improve scalability in GNN training. These settings typically involve training over locally sampled subgraphs using strategies such as random walks, clustering, or importance sampling. Our goal is to investigate how well the asynchronous Poisson-based updates perform when integrated into these localized training frameworks.

However, implementing this comparison has presented several technical challenges. Version compatibility issues between PyTorch Geometric, Python, and the available implementations of more advanced GNN models—such as GraphSAGE (\citet{graphsage}) and GAT (\citet{gat})—have made integration difficult. Initially, we intended to include these models alongside GCN in our experimental setup, as they are widely used in modern subgraph-based approaches and offer different propagation behaviors. Although our current results are based solely on GCN, extending the evaluation to include GraphSAGE and GAT remains a high priority for future work. Once these environment and implementation barriers are resolved, we expect to gain further insights into how the Poisson-based update strategy interacts with various message-passing architectures.

\section{Appendix}

\begin{figure}[H]

\centering

  \includegraphics[width=0.8\linewidth]{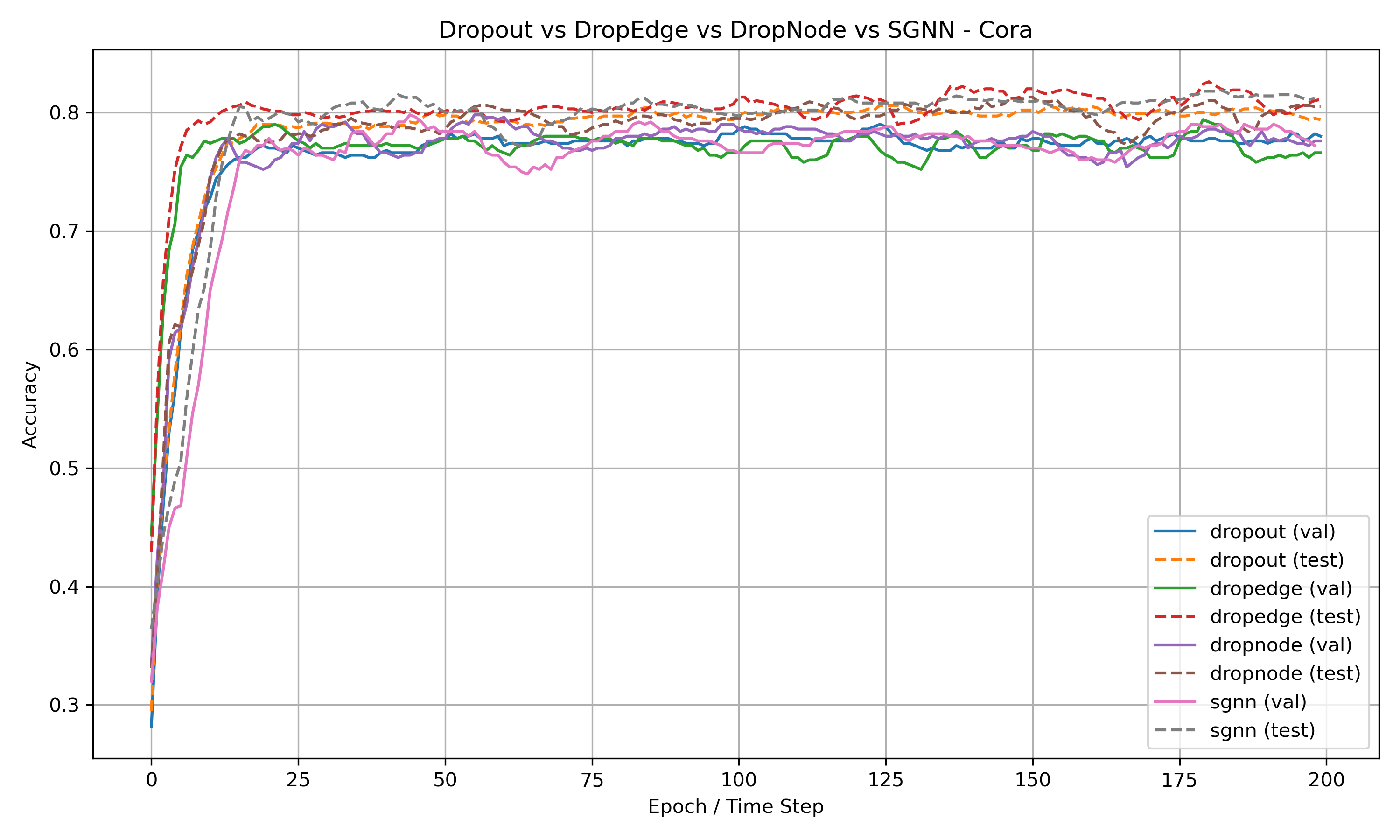}
  \caption{Cora}

\end{figure}

\begin{figure}[H]

\centering

  \includegraphics[width=0.8\linewidth]{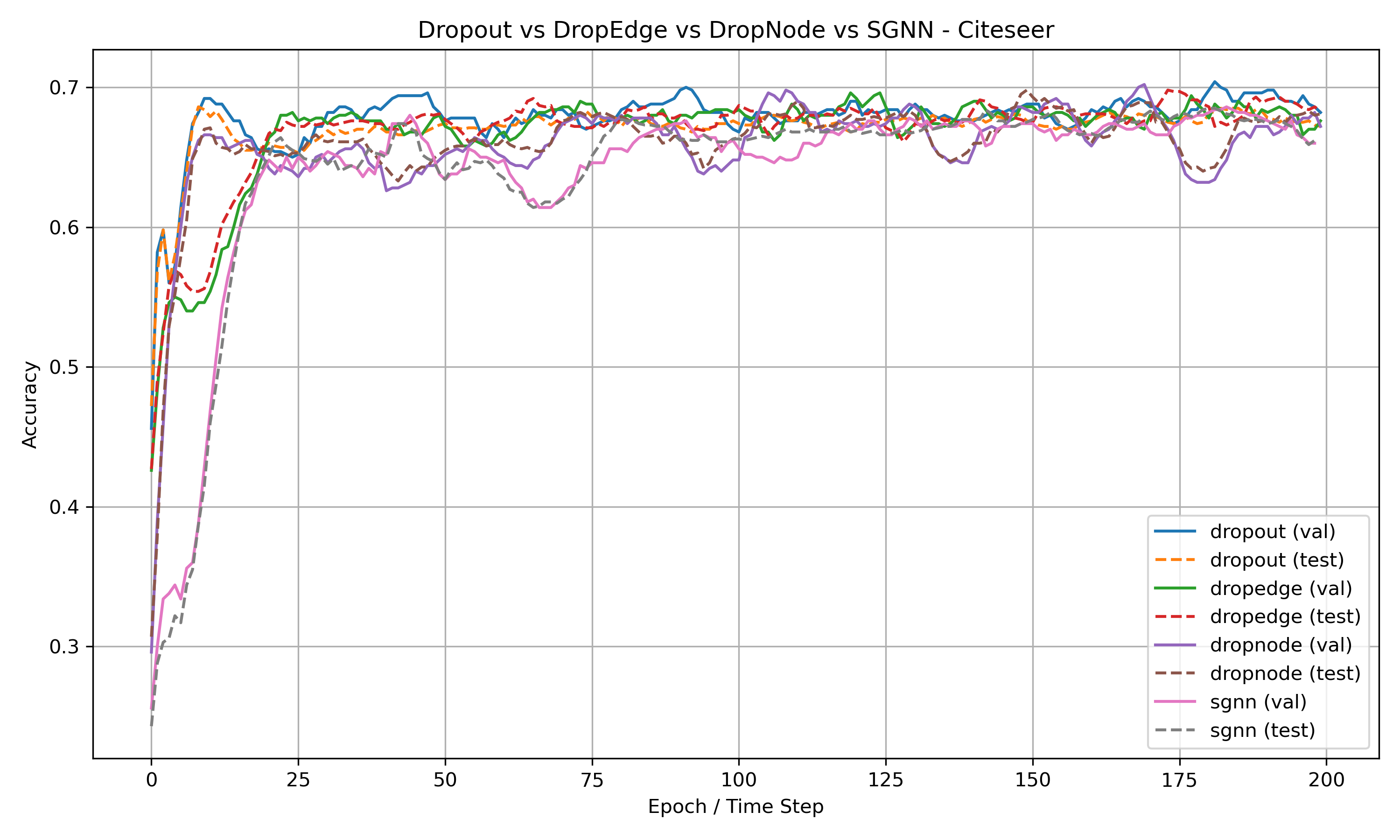}
  \caption{Citeseer}

\end{figure}

\begin{figure}[H]

\centering

  \includegraphics[width=0.8\linewidth]{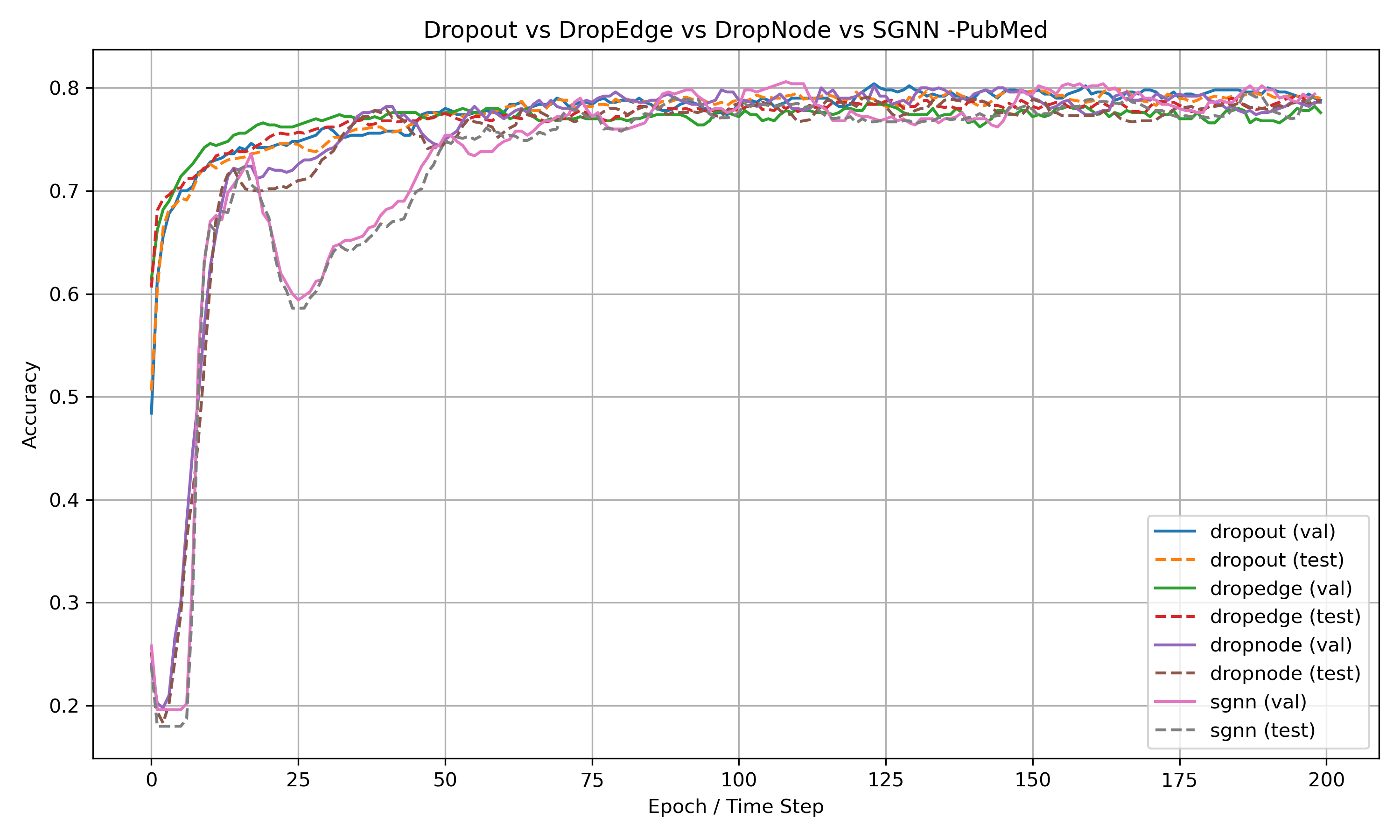}
  \caption{Pubmed}

\end{figure}

Figures show the validation and test accuracy over 200 training epochs for all datasets. While SGNN starts with slightly lower performance in early epochs compared to other methods, it exhibits more stable and improved accuracy in the later stages, especially on Pubmed.

\bibliographystyle{unsrtnat}
\bibliography{references}
\end{document}